\documentclass[11pt]{article}
\usepackage{acl2014}
\usepackage{times}
\usepackage{url}
\usepackage{latexsym}
\usepackage{graphicx} 
\usepackage{multirow}
\graphicspath{ {images/} }

\title{An Empirical Comparison of Simple Domain Adaptation Methods for \\ Neural Machine Translation}

\author{Chenhui Chu$^1$ , Raj Dabre$^2$, \and Sadao Kurohashi$^2$\\
  $^1$Japan Science and Technology Agency \\
  $^2$Graduate School of Informatics, Kyoto University \\
  {\tt chu@pa.jst.jp, raj@nlp.ist.i.kyoto-u.ac.jp, kuro@i.kyoto-u.ac.jp}} 

\date{}

\begin{document}
\maketitle
\begin{abstract}
In this paper, we propose a novel domain adaptation method named ``{\it mixed fine tuning}'' for neural machine translation (NMT). We combine two existing approaches namely {\it fine tuning} and {\it multi domain} NMT. We first train an NMT model on an out-of-domain parallel corpus, and then fine tune it on a parallel corpus which is a mix of the in-domain and out-of-domain corpora. All corpora are augmented with artificial tags to indicate specific domains. We empirically compare our proposed method against fine tuning and multi domain methods and discuss its benefits and shortcomings.
\end{abstract}

\section{Introduction}
One of the most attractive features of neural machine translation (NMT) \cite{DBLP:journals/corr/BahdanauCB14,DBLP:journals/corr/ChoMGBSB14,DBLP:journals/corr/SutskeverVL14} is that it is possible to train an end to end system without the need to deal with word alignments, translation rules and complicated decoding algorithms, which are a characteristic of statistical machine translation (SMT) systems. However, it is reported that NMT works better than SMT only when there is an abundance of parallel corpora. In the case of low resource domains, vanilla NMT is either worse than or comparable to SMT \cite{DBLP:conf/emnlp/ZophYMK16}.

Domain adaptation has been shown to be effective for low resource NMT. The conventional domain adaptation method is {\it fine tuning}, in which an out-of-domain model is further trained on in-domain data \cite{luong2015stanford,sennrich-haddow-birch:2016:P16-11,domspec,domfast}. 
However, fine tuning tends to overfit quickly due to the small size of the in-domain data.
On the other hand, {\it multi domain} NMT \cite{domcont} involves training a single NMT model for multiple domains. This method adds tags ``\textless2domain\textgreater" by modifying the parallel corpora to indicate domains without any modifications to the NMT system architecture. However, this method has not been studied for domain adaptation in particular.



Motivated by these two lines of studies, we propose a new domain adaptation method called ``{\it mixed fine tuning}," where we first train an NMT model on an out-of-domain parallel corpus, and then fine tune it on a parallel corpus that is a mix of the in-domain and out-of-domain corpora. Fine tuning on the mixed corpus instead of the in-domain corpus can address the overfitting problem. All corpora are augmented with artificial tags to indicate specific domains.
We tried two different corpora settings:
\begin{itemize}
    \item Manually created resource poor corpus: Using the NTCIR data (patent domain; resource rich) \cite{conf-ntcir-GotoCLST13} to improve the translation quality for the IWSLT data (TED talks; resource poor) \cite{cettolo2015iwslt}.
    \item Automatically extracted resource poor corpus: Using the ASPEC data (scientific domain; resource rich) \cite{NAKAZAWA16.621} to improve the translation quality for the Wiki data (resource poor). The parallel corpus of the latter domain was automatically extracted \cite{chu:2016:LREC}.
\end{itemize}
We observed that ``mixed fine tuning" works significantly better than methods that use fine tuning and domain tag based approaches separately. Our contributions are twofold:
\begin{itemize}
\item We propose a novel method that combines the best of existing approaches and have shown that it is effective.
\item To the best of our knowledge this is the first work on an empirical comparison of various domain adaptation methods.
\end{itemize}

\section{Related Work}
Besides fine tuning and multi domian NMT using tags, another direction for domain adaptation is using in-domain monolingual data. Either training an in-domain recurrent neural language (RNN) language model for the NMT decoder \cite{DBLP:journals/corr/GulcehreFXCBLBS15} or generating synthetic data by back translating target in-domain monolingual data \cite{sennrich-haddow-birch:2016:P16-11} have been studied.


\begin{figure}[t]
    \centering
  \centerline{\includegraphics[width=0.9\hsize]{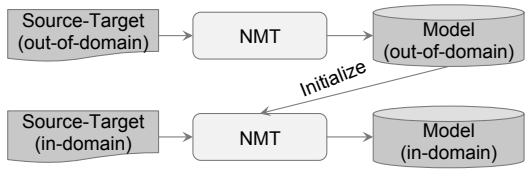}}
    \caption{Fine tuning for domain adaptation.}
    \label{finetuning}
\end{figure}

\begin{figure*}[t]
    \centering
  \centerline{\includegraphics[width=0.7\hsize]{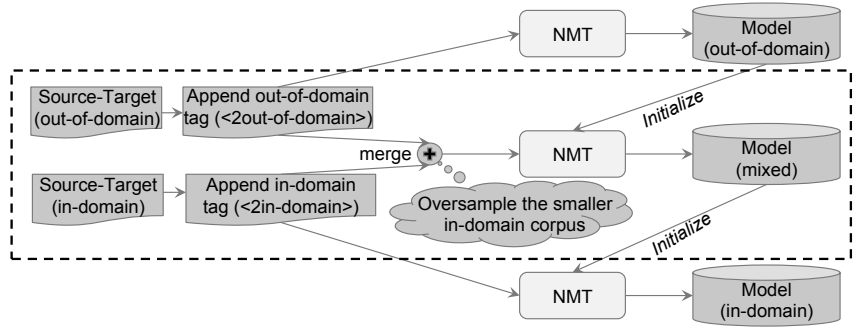}}
    \caption{Mixed fine tuning with domain tags for domain adaptation (The section in the dotted rectangle denotes the multi domain method).}
    \label{mlnmt}
\end{figure*}

\section{Methods for Comparison}
All the methods that we compare are simple and do not need any modifications to the NMT system.
\label{sec:overview}
\subsection{Fine Tuning}
{\it Fine tuning} is the conventional way for domain adaptation, and thus serves as a baseline in this study. 
In this method, we first train an NMT system on a resource 
rich out-of-domain corpus till convergence, and then fine tune its parameters on a resource poor in-domain corpus (Figure \ref{finetuning}).

\subsection{Multi Domain}
The {\it multi domain} method is originally motivated by \cite{sennrich-haddow-birch:2016:N16-1}, which 
uses tags to control the politeness of NMT translations. The overview of this method is shown in the dotted section in Figure \ref{mlnmt}. 
In this method, we simply concatenate the corpora of multiple domains with two small modifications: a. Appending the domain tag
``\textless2domain\textgreater" to the source sentences of the respective corpora.\footnote{We verified the effectiveness of
the domain tags by comparing against a setting that does not use them, see the ``w/o tags'' settings in Tables \ref{table:IWSLT2015-CE-NTCIR-CE} and
\ref{table:WIKI-CJ-ASPEC-CJ}.} 
This primes the NMT decoder to generate sentences for the specific domain. b. Oversampling the smaller corpus so that the training 
procedure pays equal attention to each domain.

We can further fine tune the multi domain model on the in-domain data, which is named as ``multi domain + fine tuning.''

\subsection{Mixed Fine Tuning}
The proposed {\it mixed fine tuning} method is a combination of the above methods (shown in Figure \ref{mlnmt}).
The training procedure is as follows:
\begin{enumerate}
\item Train an NMT model on out-of-domain data till convergence.
\item Resume training the NMT model from step 1 on a mix of in-domain and out-of-domain data (by oversampling the in-domain data) till convergence.
\end{enumerate}
By default, we utilize domain tags, but we also consider settings where we do not use them (i.e., ``w/o tags''). We can further fine tune the model 
from step 2 on the in-domain data, which is named as ``mixed fine tuning + fine tuning.''

Note that in the ``fine tuning'' method, the vocabulary obtained from the out-of-domain data is used for the 
in-domain data; while for the ``multi domain'' and ``mixed fine tuning'' methods, we use a vocabulary obtained from the 
mixed in-domain and out-of-domain data for all the training stages.

\section{Experimental Settings}
\label{sec:settings}
We conducted NMT domain adaptation experiments in two different settings as follows:

\subsection{High Quality In-domain Corpus Setting}
Chinese-to-English translation was the focus of the high quality in-domain corpus setting. We utilized the resource rich patent out-of-domain data to augment the resource poor spoken language in-domain data. The patent domain MT was conducted on the Chinese-English subtask (NTCIR-CE) of the patent MT task at the NTCIR-10 workshop\footnote{http://ntcir.nii.ac.jp/PatentMT-2/}
\cite{conf-ntcir-GotoCLST13}. The NTCIR-CE task uses 1000000, 2000, and 2000 sentences for training, development, and testing, respectively. The spoken domain MT was conducted on the Chinese-English subtask (IWSLT-CE) of the TED talk MT task at the IWSLT 2015 workshop \cite{cettolo2015iwslt}. The IWSLT-CE task contains 209,491 sentences for training. We used the dev 2010 set for development, containing 887 sentences.
We evaluated all methods on the 2010, 2011, 2012, and 2013 test sets, containing 1570, 1245, 1397, and 1261 sentences, respectively. 

\subsection{Low Quality In-domain Corpus Setting}
Chinese-to-Japanese translation was the focus of the low quality in-domain corpus setting. We utilized the resource rich scientific out-of-domain data to  augment the resource poor Wikipedia (essentially open) in-domain data. The scientific domain MT was conducted on the
Chinese-Japanese paper excerpt corpus (ASPEC-CJ)\footnote{http://lotus.kuee.kyoto-u.ac.jp/ASPEC/}
\cite{NAKAZAWA16.621}, which
is one subtask of the workshop on Asian translation (WAT)\footnote{http://orchid.kuee.kyoto-u.ac.jp/WAT/}
\cite{nakazawa-EtAl:2015:WAT}. The ASPEC-CJ task uses 672315, 2090, and 2107 sentences for training,
development, and testing, respectively. The Wikipedia domain task was conducted on a Chinese-Japanese
corpus automatically extracted from Wikipedia (WIKI-CJ) \cite{chu:2016:LREC} using the ASPEC-CJ corpus
as a seed. The WIKI-CJ task contains
136013, 198, and 198 sentences for training, development, and testing, respectively.

\subsection{MT Systems}
For NMT, we used the KyotoNMT system\footnote{https://github.com/fabiencro/knmt} \cite{cromieres-EtAl:2016:WAT2016}. The NMT training settings are the same as those of the best systems that participated in WAT 2016.
The sizes of the source and target vocabularies, the
source and target side embeddings, the hidden states, the attention mechanism hidden states, and the deep
softmax output with a 2-maxout layer were set to 32,000, 620, 1000, 1000, and 500, respectively.
We used 2-layer LSTMs for both the source and target sides. ADAM was used as the
learning algorithm, with a dropout rate of 20\% for the inter-layer dropout, and L2 regularization with a weight decay coefficient of 1e-6. The mini batch size was 64, and sentences longer than 80 tokens
were discarded. 
We early stopped the training process when we observed that the BLEU score of the development set converges. For testing, we self ensembled the three parameters of the best development loss, the best development BLEU, and the final parameters. Beam size was set to 100.

\begin{table*}
\small
\begin{center}
\begin{tabular}{@{}l|r|r|r|r|r|r@{}}\hline 
 & & \multicolumn{5}{c}{\bf IWSLT-CE} \\
\bf System & \bf NTCIR-CE &\bf test 2010 &\bf test 2011 &\bf test 2012 &\bf test 2013 &\bf average\\ \hline 
IWSLT-CE SMT & - & 12.73 & 16.27 & 14.01 & 14.67 & 14.31 \\ 
IWSLT-CE NMT & - & 6.75 & 9.08 & 9.05 & 7.29 & 7.87 \\ 
NTCIR-CE SMT & 29.54 & 3.57 & 4.70 & 4.21 & 4.74 & 4.33 \\ 
NTCIR-CE NMT & 37.11 & 2.23 & 2.83 & 2.55 & 2.85 & 2.60 \\ \hline
Fine tuning & 17.37 & 13.93 & 18.99 & 16.12 & 17.12 &  16.41 \\ 
Multi domain & 36.40 & 13.42 & 19.07 & 16.56 & 17.54 & 16.34 \\ 
Multi domain w/o tags & 37.32 & 12.57 & 17.40 & 15.02 & 15.96 & 14.97 \\ 
Multi domain + Fine tuning & 14.47 & 13.18 & 18.03 & 16.41 & 16.80 & 15.82 \\ 
{\bf Mixed fine tuning} & {37.01} & {\bf 15.04} & {\bf 20.96} & {\bf 18.77} & {\bf 18.63} & {\bf 18.01} \\ 
{Mixed fine tuning w/o tags} & {\bf 39.67} & {14.47} & {\bf 20.53} & {18.10} & {17.97} & {17.43} \\ 
{Mixed fine tuning + Fine tuning} & 32.03 & {14.40} & {19.53} & {17.65} & {17.94} & {17.11} \\ \hline
\end{tabular}
\end{center}
\caption{\label{table:IWSLT2015-CE-NTCIR-CE} Domain adaptation results (BLEU-4 scores) for IWSLT-CE using NTCIR-CE.}
\end{table*}

For performance comparison, we also conducted experiments on phrase based SMT (PBSMT).
We used the Moses PBSMT system \cite{koehn-EtAl:2007:PosterDemo}
for all of our MT experiments. For the respective tasks, we trained 5-gram language models on the target side of the training data using the KenLM toolkit\footnote{https://github.com/kpu/kenlm/}
with interpolated Kneser-Ney discounting, respectively. 
In all of our experiments, we used the GIZA++ toolkit\footnote{http://code.google.com/p/giza-pp}
for word alignment; tuning was performed by minimum error rate training \cite{och:2003:ACL},
and it was re-run for every experiment.

For both MT systems, we preprocessed the data as follows. For Chinese, we used KyotoMorph\footnote{https://bitbucket.org/msmoshen/kyotomorph-beta} for segmentation, which was trained on the CTB version 5 (CTB5) and SCTB \cite{chu-EtAl:2016:ALR12}. 
For English, we tokenized and lowercased the sentences using the {\it tokenizer.perl} script in Moses. Japanese was segmented using JUMAN\footnote{http://nlp.ist.i.kyoto-u.ac.jp/EN/index.php?JUMAN} \cite{kurohashi--EtAl:1994}. 

For NMT, we further split the words into sub-words using byte pair encoding (BPE) \cite{sennrich-haddow-birch:2016:P16-12}, which has been shown to be effective for the rare word problem
in NMT. Another motivation of using sub-words is making the different domains share more
vocabulary, which is important especially for the resource poor domain. For the Chinese-to-English tasks, we trained two BPE models on the Chinese and English vocabularies, respectively. For the Chinese-to-Japanese tasks, we trained a joint BPE model on both of the Chinese and Japanese vocabularies,
because Chinese and Japanese could share some vocabularies of Chinese characters. The number of merge
operations was set to 30,000 for all the tasks.


\section{Results}

Tables \ref{table:IWSLT2015-CE-NTCIR-CE} and \ref{table:WIKI-CJ-ASPEC-CJ} show the translation results
on the Chinese-to-English and Chinese-to-Japanese tasks, respectively. The entries with SMT and NMT are
the PBSMT and NMT systems, respectively; others
are the different methods described in Section \ref{sec:overview}.
In both tables, the numbers in bold indicate the best system and all systems that were not significantly 
different from the best system. The significance tests were performed using the bootstrap resampling method \cite{koehn:2004:EMNLP} at $p < 0.05$.


We can see that without domain adaptation, the SMT systems perform significantly better than the 
NMT system on the resource poor domains, i.e., IWSLT-CE and WIKI-CJ; while on the resource rich domains,
i.e., NTCIR-CE and ASPEC-CJ, NMT outperforms SMT. 
Directly using the SMT/NMT models trained on the out-of-domain data to translate the 
in-domain data shows bad performance. With our proposed ``Mixed fine tuning" domain adaptation method,
NMT significantly outperforms SMT on the in-domain tasks.


\begin{table}
\small
\begin{center}
\begin{tabular}{@{}l|r|r@{}}\hline 
\bf System &\bf ASPEC-CJ &\bf WIKI-CJ \\ \hline 
WIKI-CJ SMT & - & 36.83 \\ 
WIKI-CJ NMT & - & 18.29 \\ 
ASPEC-CJ SMT & 36.39 & 17.43 \\ 
ASPEC-CJ NMT & \bf 42.92 & 20.01 \\ \hline
Fine tuning & 22.10 & \bf 37.66 \\ 
Multi domain & 42.52 & 35.79 \\ 
Multi domain w/o tags & 40.78 & 33.74 \\ 
Multi domain + Fine tuning & 22.78 & 34.61 \\ 
{\bf Mixed fine tuning} & {\bf 42.56} & {\bf 37.57} \\ 
{Mixed fine tuning w/o tags} & {41.86} & {\bf 37.23} \\ 
{Mixed fine tuning + Fine tuning} & 31.63 & {\bf 37.77} \\ \hline
\end{tabular}
\end{center}
\caption{\label{table:WIKI-CJ-ASPEC-CJ} Domain adaptation results (BLEU-4 scores) for WIKI-CJ using ASPEC-CJ.}
\end{table}

Comparing different domain adaptation methods, ``Mixed fine tuning'' shows the best performance. 
We believe the reason for this is that ``Mixed fine tuning'' can address the over-fitting problem of ``Fine tuning.'' 
We observed that while ``Fine tuning'' overfits quickly after only 1 epoch of training, ``Mixed fine tuning'' only 
slightly overfits until covergence. In addition, ``Mixed fine tuning'' does not worsen the quality
of out-of-domain translations, while ``Fine tuning'' and ``Multi domain'' do. One shortcoming of ``Mixed fine tuning'' is that compared 
to ``fine tuning,'' it took a longer time for the fine tuning process, as the time until convergence is essentially 
proportional to the size of the data used for fine tuning. 

``Multi domain'' performs either as well as (IWSLT-CE) 
or worse than (WIKI-CJ) ``Fine tuning,'' but ``Mixed fine tuning'' performs either significantly better than (IWSLT-CE) or is
comparable to (WIKI-CJ) ``Fine tuning.'' We believe the performance difference between the two tasks is due to their 
unique characteristics. As WIKI-CJ data is of relatively poorer quality, mixing it with out-of-domain data does not 
have the same level of positive effects as those obtained by the IWSLT-CE data. 

The domain tags are helpful for both ``Multi domain'' and ``Mixed fine tuning.'' Essentially, further fine tuning 
on in-domain data does not help for both ``Multi domain'' and ``Mixed fine tuning.'' We believe the reason 
for this is that the ``Multi domain'' and ``Mixed fine tuning'' methods already utilize the in-domain data used 
for fine tuning. 


\section{Conclusion}
In this paper, we proposed a novel domain adaptation method named ``mixed fine tuning'' for NMT. We empirically compared our proposed method against fine tuning and multi domain methods, and have shown that it is effective but is sensitive to the quality of the in-domain data used.

In the future, we plan to incorporate an RNN model into our current architecture to leverage abundant in-domain monolingual corpora. We also plan on exploring the effects of synthetic data by back translating large in-domain monolingual corpora.
\bibliography{acl2017}
\bibliographystyle{acl}

\end{document}